# annbatch unlocks terabyte-scale training of biological data in anndata

Ilan Gold[1 #], Felix Fischer[1,2,3 #], Lucas Arnoldt[1,4], F. Alexander Wolf[5], Fabian J. Theis[1,2,4]

[1]Department of Computational Health, Institute of Computational Biology, Helmholtz Munich, Germany
[2]School of Computing, Information and Technology, Technical University of Munich, Munich, Germany
[3]Munich Center for Machine Learning (MCML)
[4]TUM School of Life Sciences Weihenstephan, Technical University of Munich, Munich, Germany
[5]Lamin Labs

# Equal contribution

## Abstract

The scale of biological datasets now routinely exceeds system memory, making data access rather than model computation the primary bottleneck in training machine-learning models. This bottleneck is particularly acute in biology, where widely used community data formats must support heterogeneous metadata, sparse and dense assays, and downstream analysis within established computational ecosystems. Here we present annbatch, a mini-batch loader native to anndata that enables out-of-core training directly on disk-backed datasets. Across single-cell transcriptomics, microscopy, and whole-genome sequencing benchmarks, annbatch increases loading throughput by up to an order of magnitude and shortens training from days to hours, while remaining fully compatible with the scverse ecosystem. Annbatch establishes a practical data-loading infrastructure for scalable biological AI, enabling the use of increasingly large and diverse datasets without abandoning standard biological data formats. Github: https://github.com/scverse/annbatch

## Main

Contrary to prevailing intuition, state-of-the-art (SoTA) deep learning models in many domains of biology are often data loading bound, meaning that prolonged fitting times often stem from inefficient data retrieval rather than model complexity due to their smaller model size[1,2]. Addressing these data loading limitations offers the potential to reduce training times by several orders of magnitude, effectively shifting the bottleneck back to the hardware's actual processing power.

To address this issue and achieve the high-performance data loading speeds required by modern GPUs, one needs to switch from purely random disk access to pseudo-random access that reads data in chunks, resulting in significantly faster read speeds. Such approaches have already been popularized in the wider machine learning community, exemplified by frameworks such as NVIDIA Merlin[3,4], Uber Petastorm[5], and WebDataset[6]. However, applying such frameworks to current data formats popular in the biomedical community – most notably anndata[7] – requires file-format conversion, meaning the output formats from these frameworks do not integrate with widely used community tools. Furthermore, saturating a modern GPU requires reading considerably larger chunks than those recommended by existing tools, such as scDataset[8]. To enable reading larger sequential chunks, it is required to pre-shuffle the on-disk data to guarantee batch diversity. However, tools for efficiently pre-shuffling large data collections that do not fit in memory do not exist yet in the single-cell community.

To solve these issues, we introduce annbatch, an easy-to-use framework that fully integrates with anndata and that provides utility functions from pre-shuffling the input data to ready-to-use, high-performance data loaders. This closes a critical gap that has hindered the adoption of highly performant data loading solutions to date.

# Results

## Annbatch: A high-performance data loader for the anndata file format

Annbatch provides an end-to-end high-throughput data loading framework that integrates with the scverse ecosystem (Fig. 1a). Such an end-to-end solution does not exist yet: While existing solutions often rely on data formats that hinder interoperability (such as BioNemo[9] or SLAF[10]) or suffer from below-GPU-saturation data loading speeds (scDataset[8] or MappedCollection[11]) (Fig. 1b), annbatch addresses both limitations. This integration is critical in single-cell biology: unlike in traditional machine learning domains, the lack of a definitive ground truth requires users to validate model outputs using established ecosystem tools. Annbatch enables this rigorous downstream analysis without sacrificing computational efficiency by remaining firmly within the scverse[12] ecosystem - inputs and outputs are simply anndata files.

To accomplish this task, we provide two core functionalities in the annbatch library - a to-date novel pre-shuffler of on-disk anndata files and a highly performant data loader (for the pre-shuffled anndata files). Preshuffling data ensures that the loader's pseudo-random chunked disk-access pattern can fetch large enough contiguous regions of the on-disk data for fast training without compromising the intra- and inter-batch randomness crucial to minibatch learning[13]. With a flexible API and low memory requirements enabled by anndata's lazy loading, preshuffling is efficient, configurable, and can be run from an HPC environment to a laptop without leaving the anndata file format. The data loader then leverages the pre-shuffled nature of the on-disk data to fetch large, randomized, contiguous blocks of observations rather than individual observations or small contiguous blocks (see Methods), thereby fully exploiting sequential I/O. The loader implementation relies on zarrs-python, a bridge between Rust-based zarrs and Python-based zarr-python that we developed as well. We further increase loading speeds through state-of-the-art techniques such as custom uneven-slice indexing of Zarr arrays, direct I/O, a Linux feature primarily used in high-performance database systems, reading into pinned memory, and lastly, GPU acceleration. Crucially, our data loader is independent of a specific deep learning framework.

In summary, annbatch addresses the primary technical bottlenecks that currently impede the adoption of high-performance data loaders in the anndata ecosystem. By incorporating essential features such as optimized pre-shuffling logic while maintaining full interoperability with the anndata ecosystem, annbatch bridges the gap between high data loading throughput and user accessibility. These advancements establish a streamlined path for integrating high-throughput data loading protocols into all workflows that can be mapped to the anndata format.

Figure 1: annbatch is a flexible toolbox for loading on-disk anndata files that alleviates I/O bottlenecks in minibatch training

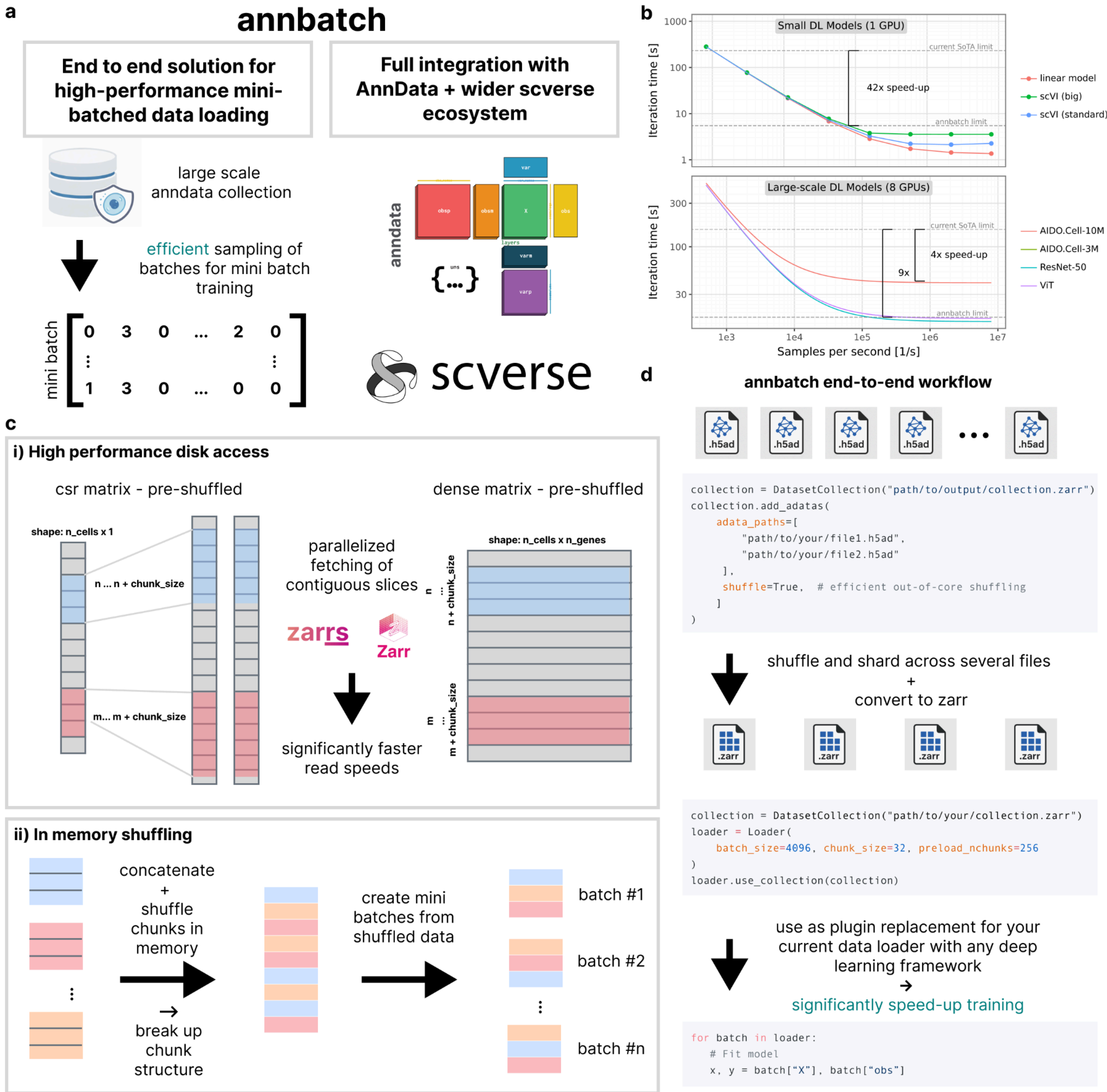

**a)** Annbatch provides an end-to-end data loading solution for mini-batch training and fully integrates with the anndata and wider scverse ecosystem. **b)** Visualization of how data loading speed affects model training time. The plot shows fit times for standard, smaller deep/machine learning models (scVI and logistic regression) on a single GPU and large-scale deep learning models (AIDO.Cell model with 3 and 10 million parameters, a ResNet model and a vision transformers model (ViT)). The dashed lines indicate the throughput limits of current state-of-the-art modes (MappedCollection[11]) and annbatch, respectively. Potential speed-ups are indicated by the respective brackets. **c)** Schematic sketch of the key improvements of annbatch: i) contiguous disk access, meaning data gets fetched as randomly selected contiguous chunks instead of individual elements; *ii)* in-memory shuffling, meaning the randomly selected contiguous chunks from step *i)* get shuffled in memory before being grouped into mini-batches. **d)** Overview of the most important processing steps in the annbatch API.

## Zarr-based annbatch speeds up model fits by orders of magnitudes

Annbatch demonstrates a substantial increase in data throughput compared to existing frameworks. To evaluate real-world utility, we benchmarked iteration times on the Tahoe100M dataset (~100 million cells).

Under default settings, annbatch achieves ~35,000 samples per second (samples/sec), whereas scDataset and MappedCollection reach only ~1,500 and ~850 samples/sec, respectively (Fig. 2a). These speed improvements yield a nearly 40-fold acceleration in model fitting compared to MappedCollection (Fig. 2b). Note: These benchmarks were performed on an Amazon Web Services (AWS) ml.g4dn.4xlarge instance with standard Elastic Block Storage (EBS) to ensure reproducible baseline conditions. On AWS the disk speed is limited to 125MB/s resutling in slower loading speeds and no benefit from GPU acceleration. With a more performant storage/file system, data loading is faster (see de.NBI results in Fig. 2a). In more concrete terms, annbatch processes one epoch in approximately 17 minutes using GPU acceleration, while scDataset requires ~7.5 hours and MappedCollection ~21 hours (Fig. 2b). These efficiency improvements yield significant GPU time savings per epoch. Notably, the ~5h pre-shuffling overhead required by annbatch, which does not require a GPU, is already fully amortized within the first training epoch compared to MappedCollection and scDataset. Moreover, even though the anndata format is standardized, anndata files are often heterogeneous, e.g. raw counts might be stored under the .X or the .raw.X attribute, batch labels might be stored under different column names, etc. Curating such (often large) anndata collections usually thus already entails writing new data to disk, which makes pre-shuffling simply a final step in the pipeline rather than a “duplication” of data. Annbatch’s flexible API fully supports this data curation by providing a customizable loading function and warnings on “mismatched” items, such as columns present in only one object.

A big contributor to these performance improvements is the adoption of Zarr as the storage back-end, which outperforms the standard h5ad format thanks to the zarrs-python package. Transitioning to Zarr reduced pre-shuffling overhead from ~12 hours to ~5.5 hours (3.5 hours if the input files are in Zarr format as well) (Fig. 2d) and increased throughput from ~20,000 to ~54,000 samples/sec (Fig. 2d). These gains were achieved while maintaining about the same disk usage (Fig. 2d) and are further increased when factoring in data transfer to the GPU for which annbatch provides optimized handling (Fig. 2a).

Figure 2: annbatch achieves high-performance data loading speeds across a wide range of biological applications

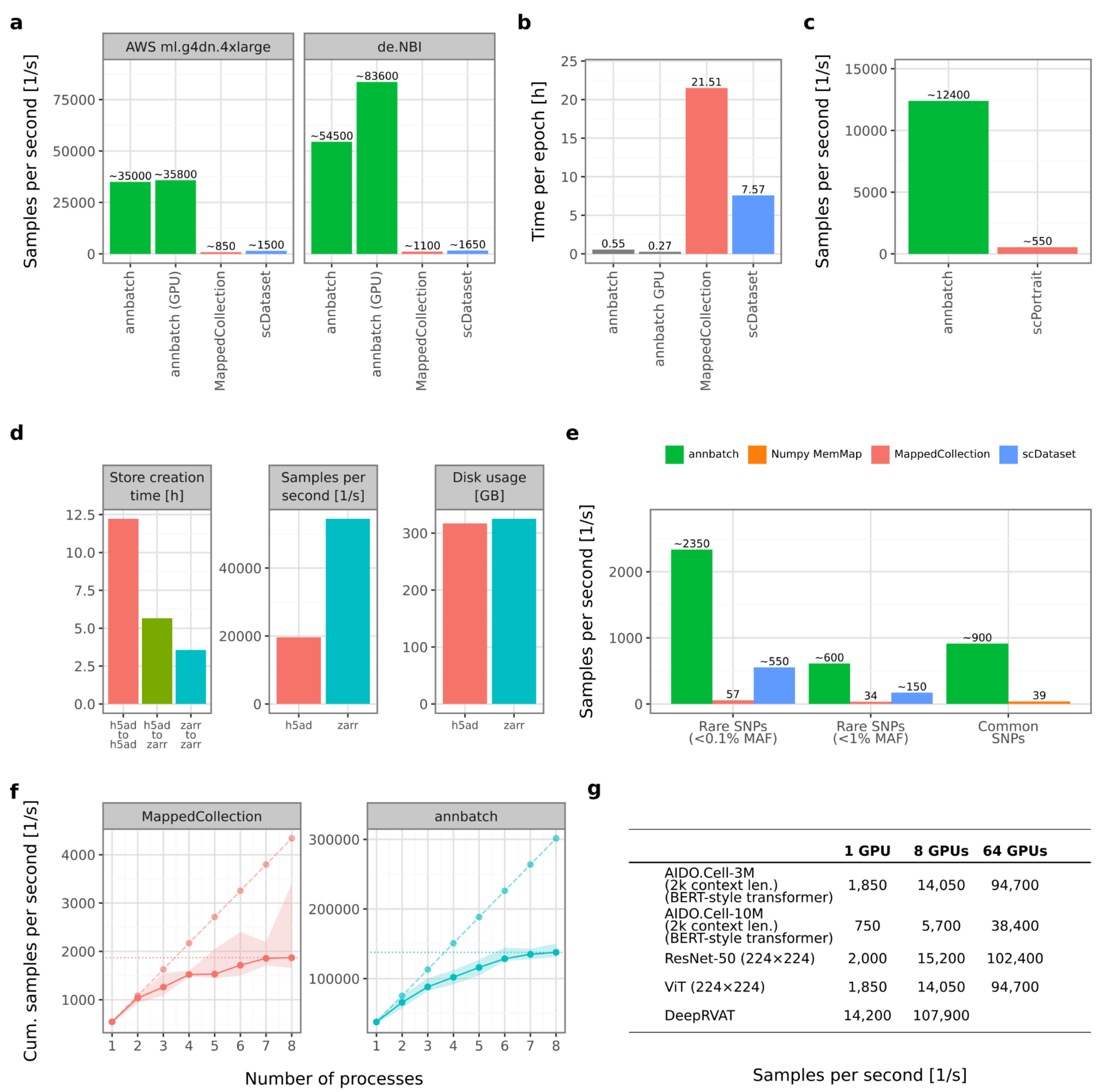

| | 1 GPU | 8 GPUs | 64 GPUs |
|---|---|---|---|
| AIDO.Cell-3M (2k context len.) (BERT-style transformer) | 1,850 | 14,050 | 94,700 |
| AIDO.Cell-10M (2k context len.) (BERT-style transformer) | 750 | 5,700 | 38,400 |
| ResNet-50 (224×224) | 2,000 | 15,200 | 102,400 |
| ViT (224×224) | 1,850 | 14,050 | 94,700 |
| DeepRVAT | 14,200 | 107,900 | |

**a)** Data loading speeds on the Tahoe-100M dataset (scRNA-seq data) on an AWS ml.g4dn.4xlarge and de.NBI GPU T4 medium instance. **b)** Time to iterate over one epoch of the Tahoe-100M dataset. **c)** Data loading speeds for single-cell microscopy images (size: 128x128x5). **d)** *i)* Store creation time for the Tahoe-100M dataset using Zarr and h5ad storage formats. *ii)* Data loading speeds comparing data loading speeds for chunked data loading (chunk/fetch settings matched between annbatch with zarr and scDataset with h5ad). *iii)* Disk usage for the Tahoe-100M for Zarr and h5ad storage formats. **e)** Data loading speeds for whole-genome sequencing data benchmarked on the 1000 Genomes GRCh38 dataset expanded to 500,000 individuals, across three variant regimes: rare variants at MAF < 0.01% (~72M SNPs, sparse) and MAF < 0.1% (~90M SNPs, sparse), and common variants at MAF ≥ 1% (1.2M SNPs, dense). MappedCollection and scDataset were omitted from the MAF < 0.1% comparison due to prohibitively low loading speeds. **f)** Cumulative yielded samples per second (aggregated across all data loading processes) plotted against the number of simultaneously running data loading processes on a single NVIDIA HGX-H100 unit. **g)** Reference throughput numbers for exemplary large-scale deep learning models on a NVIDIA H100 GPU. Throughput for 8 and 64 GPUs is extrapolated using 95% and 80% scaling efficiency, respectively.

# Annbatch speeds up model fitting across a range of biological data types

Although annbatch was developed in the context of single-cell omics, its design targets three fundamental data loading regimes that dominate modern biological machine learning: very large numbers of samples, very high feature dimensionality, and large per-sample data payloads. While the Tahoe-100M benchmark stresses the first regime, annbatch is equally applicable to modalities that challenge the latter two. By natively supporting the anndata data format backed by Zarr, annbatch provides a robust, high-performance foundation for large-scale data loading, ensuring seamless generalizability across a broad spectrum of biological modalities.

## High-performance data loading for neural-network-based genetics models

To demonstrate applicability beyond single-cell data, we benchmarked annbatch on whole-genome sequencing data. Training throughput in genetics is often limited by data loading rather than model complexity. For example, DeepRVAT, a deep-set model for rare-variant association testing with relatively few trainable parameters (Fig. 2g), already employs a custom Zarr-based data loading solution for scalability[2] and would additionally benefit from annbatch's integrated pre-shuffling. As the field moves from exome to whole-genome sequencing, efficient general-purpose data loaders become increasingly essential.

We evaluated annbatch on the publicly available 1000 Genomes GRCh38 dataset[14], expanded by duplication to 500,000 individuals to approximate the scale of restricted-access cohorts such as the UK Biobank[15] and All of Us Research Program[16]. 1000 Genomes was chosen because it is publicly available and approximates relevant sparsity patterns. Conversion of raw files to the anndata Zarr format was performed using GenoAnnData[17], making it straightforward to create whole-genome-scale anndata stores for any cohort.

For rare variants stored in sparse anndata format, which avoids the need for separate variant-annotation files required by alternative representations[2], annbatch achieves ~41-fold and ~2.8-fold speed-ups over MappedCollection and scDataset at MAF < 0.01% (~72M SNPs), and ~18-fold and ~3.2-fold at MAF < 0.1% (~90M SNPs) (Fig. 2e). The higher absolute throughput at MAF < 0.01% reflects that the MAF < 0.1% dataset is approximately four times larger, as the additional SNPs are substantially less sparse. For common variants (1.2M SNPs, matching HapMap3-style selection used in recent polygenic modeling work[18,19], annbatch achieves a ~23-fold speed-up over a numpy memmap (Fig. 2e), an uncompressed disk-backed baseline that becomes storage-infeasible as models incorporate larger variant sets or longer context windows. MappedCollection and scDataset were omitted from the common SNP comparison as their loading speeds were too low to benchmark within a reasonable time.

### High-throughput loading of single-cell microscopy images via annbatch

To showcase a large per-sample payload, we can consider applying annbatch to cropped single-cell microscopy images as modeled by scportrait[20]. The anndata-backed format developed by scportrait provides an efficient way to store the microscopy images in the .obsm attribute. Similar to the scRNA-seq example, annbatch can provide substantial speed-ups over random indexing into h5ad files – approximately a 22x speed-up (Fig. 2c).

## Distributed model training with annbatch

In a distributed training regime, data ingestion frequently becomes a critical bottleneck for workloads that remain compute-bound in single-accelerator configurations (Fig. 2f). While instantiating parallel data loaders across multiple nodes increases aggregate throughput, the scaling remains fundamentally sublinear due to shared resource contention. Consequently, I/O performance often fails to match the near-linear scaling of model throughput across distributed compute clusters (Fig. 2g). By substantially increasing baseline data loading efficiency, annbatch maintains sufficient throughput to saturate high-performance distributed architectures, effectively decoupling training performance from I/O constraints. Concretely, both vision transformers common in imaging and BERT-style transformers are often trained in multi-GPU settings due to their large number of parameters[20].

# Discussion

Annbatch addresses a critical infrastructure gap in the scverse ecosystem, extending the use of anndata, the foundational data structure, from a static storage format to a high-performance data loading solution for deep learning. By providing a streamlined, end-to-end data loading architecture based on the Zarr storage format, we demonstrate that I/O-related overhead — previously a prohibitive factor in model training — can be efficiently mitigated, potentially reducing model fitting times by several orders of magnitude. Crucially, this performance gain does not come at the cost of interoperability since the on-disk data are anndata files and thus compatible with the scverse ecosystem at large. We further show through our results on microscopy and genetics data that the widespread adoption of both the anndata and Zarr formats makes it simple to use our data loader in a variety of biomedical contexts.

However, annbatch currently has some limitations that pose further engineering challenges that need to be addressed in the future. First, consecutive reads make it difficult to implement arbitrary sampling strategies, most notably weighted sampling. To address this, data needs to be written on disk in a way that respects the sampling strategy. Future versions of annbatch will provide utility functions to help users handle this without writing custom code. Second, annbatch does not directly support multi-modal data. Such functionality can be extended in the future as well; however, we would argue that this is not as high a priority at the moment, as multi-modal data is often not available at scale. Lastly, we would like to address the preshuffling overhead again. Here, it's important to note that purely random access is often very inefficient, leading to slow data loading speeds that can be a severe bottleneck, especially for smaller deep learning models. Consequently, the pre-shuffling overhead is often already amortized after one or two epochs.

Furthermore, a deep integration with the Zarr file format enables annbatch to leverage many features in the continuously growing Zarr ecosystem. For example, icechunk provides transaction-safe, versioned storage of data that is consumable by Zarr implementations in both Rust and Python, thus enabling collaborative, versioned model training and inference[20]. Furthermore, our implementation is fully async-compatible thanks to zarr-python, thus enabling cloud-based training for appropriate workflows (although we recommend using local, disk-backed data when possible due to likely network bottlenecks). Lastly, given the speed improvements for preshuffling of starting with anndata stored as zarr, we thus hope this paper serves as a motivation for the community to store anndata files in zarr instead of hdf5. To this end, scverse hopes to upstream zarr-backed anndata into tools like cyto[21] and alevin-fry[22] so that quantified reads are immediately stored in this format instead of matrix market formats or hdf5, thus benefitting all downstream I/O tasks.

As the community moves toward ever larger datasets, the ability to efficiently train such models through optimized, reusable data ingestion will be indispensable. We anticipate that annbatch will serve as a foundational tool for the next generation of scalable biological discovery, enabling researchers to leverage increasingly massive datasets without either the historical constraints of storage-related bottlenecks or the need to write custom data loading solutions.

# Methods

## Annbatch: Workflow and main components

The annbatch library has two primary components: a pre-shuffler for writing out shuffled data to disk, and a data loader for loading the pre-shuffled data from disk in mini-batches.

### Preshuffler

The input to the preshuffler is a list of file paths pointing to different anndata files on disk. First, these files are loaded and concatenated (by default lazily - meaning the data does not get loaded into system memory). Next, to shuffle the concatenated AnnData without fully loading it into memory, the subset is read into memory by a list of random contiguous chunks i.e., the anndata object is indexed along its obs axis by a randomized list of contiguous chunks of a user-defined size (Note: the bigger the size of the contiguous chunks the faster the processing, but at the same time randomness decreases). The in-memory subset is then shuffled and written to disk as a Zarr-backed anndata. The process is repeated until all of the input data has been loaded, shuffled, and written to disk. Note that when the data is loaded lazily, the subsetting has memory requirements of O(m), where m is the user-defined amount of data to be shuffled before being written to disk, instead of O(N), where N is the total number of obs in all of the on-disk anndata objects. The preshuffled data allows the data loader to load larger, contiguous portions of on-disk data without sacrificing batch randomness. This stands in contrast to tools like scDataset[19] and cellarium-ml[23], which both implement similar pseudo-random chunked access to annbatch, but do not offer a preshuffling component.

### Data Loader

The data loader works in a similar fashion, reading many disparate chunks from the input anndata files into memory and then shuffling them. However, the on-disk anndata is now in the Zarr format for which we have written optimized I/O code. Once in RAM, asynchronous GPU transfer (via direct-to-pinned-memory reading) and acceleration can be applied to shuffling and yielding batches (instead of writing to disk as with the preshuffler).

## Importance of representative benchmarking scenarios

Benchmarking data loaders on datasets that exceed system memory capacity is essential to avoid the confounding effects of operating system (OS) caching. For example, evaluating the BioNemo[9] loader on a 4.7 million cell subset — which fits entirely within RAM — showed a throughput increase from ~2,500 samples/sec in the first epoch to ~110,000 samples/sec in the second as the data became resident in memory. Because such performance gains are not representative of large-scale datasets where in-memory residency is infeasible, annbatch was evaluated under conditions that accurately reflect real-world, out-of-core data handling.

Furthermore, as data loading throughput is intrinsically linked to hardware specifications, we standardized our primary benchmarks (Fig. 2a) on an AWS ml.g4dn.4xlarge instance to ensure comparability. Specifically, executing our benchmarks on high-performance computing (HPC) infrastructure, specifically the hardware hosted by the German network of bioinformatics infrastructure (de.NBI), yielded approximately double the data loading throughput (Fig. 2a), highlighting the scalability of the framework across different compute environments due to AWS's disk speed rate limiting.

Finally, achieving peak GPU utilization requires optimizing batch sizes (Supp. Fig. 1a) (Note: default batch size settings in e.g. scVI tools are often way too small to fully utilize a modern GPU). Increasing batch sizes significantly enhances model fit times. Notably, with an increased batch size, the learning rate needs to be scaled accordingly.

## Annbatch: Benchmarks

### Single-cell RNA-seq benchmarks (Tahoe-100M)

We ran all our benchmarks on the full 100 million cells of the Tahoe100M dataset. The data loading speed benchmarks were run either on an Amazon Web Services ml.g4dn.4xlarge instance (Fig. 2a,b) or a de.NBI (German network for bioinformatics infrastructure) GPU T4 medium instance (16 vCPUs + 64GB memory). The Zarr versus h5ad comparison (Fig. 2d,e,f) and the scaling benchmarks with respect to the number of processes (Fig. 2g) were run on the AI partition of the Leibniz Rechenzentrum (LRZ), as these benchmarks required more CPU cores.

### Single-cell microscopy image benchmarks

All benchmarks for the single-cell microscopy images were run on a de.NBI GPU T4 medium instance (16 vCPUs + 64GB memory).

### 1000Genomes benchmark dataset and preprocessing

We evaluated annbatch on the 1000 Genomes Project GRCh38 release (3,202 individuals)[14]. Per-chromosome VCF files were converted to PGEN format using PLINK2 (v2.0), applying biallelic SNP filtering (--snps-only just-acgt, --max-alleles 2) and standardized variant IDs. Chromosomes 1–22 were merged into a single PGEN file. Three variant subsets were extracted: rare variants at MAF < 0.01% (~72M SNPs) and MAF < 0.1% (~90M SNPs), and common variants at MAF ≥ 1% (subset to 1.2M SNPs, matching the HapMap3-based selection used in Kelemen et al.[12], as opposed to whole-exome-derived sets of up to 7M variants[11]). Genotype matrices were transformed into AnnData Zarr format using cellink v0.0.1, preserving sample-wise observations and variant-wise features while enabling chunked, out-of-core access. Rare-variant datasets were stored as sparse (CSR) arrays; common variants were stored dense. To reach a realistic cohort scale, the 3,202-individual dataset was replicated by duplication to 500,000 individuals; 1000 Genomes was chosen over restricted-access cohorts (UK Biobank, All of Us) as it is publicly available while approximating relevant sparsity patterns. Shuffled Zarr stores were generated using annbatch.DatasetCollection.add_adatas() with format-appropriate chunking parameters (sparse: chunk size $2^{21}$–$2^{23}$, shard size $2^{19}$–$2^{21} \times 512$; dense: chunk size 4, shard size 512). The common-variant memmap baseline stores data as uncompressed uint8 arrays on disk and accesses them out-of-core under identical memory constraints. All benchmarks were run on a Helmholtz Munich HPC node (Intel Xeon Gold 6142M, 16 vCPUs, 64 GB RAM, Lustre parallel filesystem).

## Reference models for data loading speed

### ResNet-50

We used the standard PyTorch ResNet-50 implementation. Benchmarks were run on a single H100 GPU. We used the standard ImageNet input size of 224 x 224 x 3.
https://docs.pytorch.org/vision/main/models/generated/torchvision.models.resnet50.html

### AIDO.Cell models

As transformer reference models, we used the BERT-style transformer models from Genbio AI AIDO.Cell[24]. We used two models, one with 3 million parameters and one with 10 million parameters. Both models were used with a context length of 2000 tokens. Performance was evaluated on an Nvidia H100 GPU.
3 million parameter model: https://huggingface.co/genbio-ai/AIDO.Cell-3M
10 million parameter model: https://huggingface.co/genbio-ai/AIDO.Cell-10M

### Reference models for genetics benchmarks

To contextualize the importance of data loading speed in genetic workflows, we reference three representative neural network architectures. DeepRVAT[2] is a deep-set network that aggregates rare variant annotations into a trait-agnostic gene impairment score; the gene impairment module consists of two small multilayer perceptrons (embedding network $\varphi$ and scoring network $\rho$, with hidden layer widths of 20 and 10, respectively), making it computationally lightweight despite operating on exome-scale variant sets. GenNet[18] constructs sparse, biologically interpretable networks by using prior knowledge (gene, pathway, and tissue annotations) to define only meaningful connections between millions of SNPs and downstream biological entities; the resulting architectures are memory-efficient and trainable on a single consumer GPU. Kelemen et al.[19] evaluated fully connected multi-layer perceptrons with three hidden layers (at most 100–50–25 neurons) trained on up to ~1.2M HapMap3-derived SNPs from the UK Biobank to assess nonlinear contributions to polygenic scores. All three architectures are characterized by high input dimensionality relative to the number of trainable parameters, a regime in which the forward pass is fast, and data ingestion dominates total training time.

### DeepRVAT data storage format

DeepRVAT stores rare variant data in a custom sparse format in HDF5 (v1.10.6), in which all variants per individual are stored as a variable-length array padded to the maximum number of rare variants observed in any individual, with a separate file containing variant annotations. This format requires post-processing and multi-file handling. In contrast, annbatch uses a sparse AnnData Zarr format that stores all individuals and variants in a single file with sample-wise observations and variant-wise features, eliminating the need for separate annotation files and enabling direct integration with the scverse ecosystem. For the UK Biobank 200k WES dataset, the DeepRVAT HDF5 store required approximately 100 GB of storage, compared to multiple terabytes for the original pVCF files[2].

## Vision Transformer

As a vision transformer (ViT) reference model, we used the standard PyTorch vit_b_16 reference model. Again, we used the standard ImageNet image size of 224 x 224 x 3. Performance was evaluated on an Nvidia H100 GPU.
https://docs.pytorch.org/vision/main/models/generated/torchvision.models.vit_b_16.html#torchvision.models.vit_b_16

# Acknowledgements

We thank Tim Treis for recognizing the scportrait use-case. We thank Lachlan Deakin from the Australian National University, specifically, for his tireless effort supporting the zarrs ecosystem (including zarrs-python), as well as all open source contributors to the zarr-python library, without whom this effort would not be possible.

This work was supported by the de.NBI Cloud within the German Network for Bioinformatics Infrastructure (de.NBI), ELIXIR-DE (Forschungszentrum Jülich and W-de.NBI-001, W-de.NBI-004, W-de.NBI-008, W-de.NBI-010, W-de.NBI-013, W-de.NBI-014, W-de.NBI-016, W-de.NBI-022), the European Union (ERC, DeepCell - 101054957), the Wellcome Trust, Kavli Foundation and grant reference 313280/Z/24/Z from the Chan Zuckerberg Initiative Foundation, the German Federal Ministry of Research, Technology and Space (BMFTR) under grant no. 01IS18053A, and the Helmholtz Association, as part of the joint research school Munich School for Data Science (MUDS).

# Author Contributions

F.F. and I.G. jointly conceptualized the project. I.G. and F.F. lead the software development efforts. F.F., I.G., and L.A. ran the associated benchmarks in the paper. F.F., I.G., and L.A. wrote the original manuscript draft with input from F.T. F.A.W. and F.T. acquired funding, provided resources, supervised the project, and reviewed and edited the original draft. All authors read and approved the final manuscript.

# Competing Interests Statement

F.F. consults for Lamin Labs. F.A.W. is the CEO at Lamin Labs and has ownership interest in Cellarity & Retro. F.J.T. consults for Immunai, CytoReason, BioTuring, Genbio and Valinor Industries, and has ownership interest in RN.AI Therapeutics, Dermagnostix, and Cellarity. The remaining authors declare no conflicts of interest.

# Data Availability

All datasets used in this paper are publicly available at the following links:

- Tahoe100M dataset:
  https://lamin.ai/laminlabs/arrayloader-benchmarks/collection/eAgoduHMxuDs5Wem
- Single-cell microscopy images:
  https://github.com/MannLabs/scPortrait_manuscript/blob/545c0c34593d04ddb54235fbd5c42538f9aa0a60/figure_data/download_data.sh#L13-L17
- Genomics experiments:
  https://www.internationalgenome.org/data-portal/data-collection/1000genomes

# Code Availability

annbatch package:

- Package: https://github.com/scverse/annbatch
- Documentation: https://annbatch.readthedocs.io/en/stable/

Benchmarks:

- https://github.com/theislab/annbatch_paper

BioNemo benchmarks:

- https://github.com/felix0097/BioNemoSCDL_benchmark/blob/main/bionemo_benchmark.ipynb

Supp. Fig 1: Batch size versus total fit time.

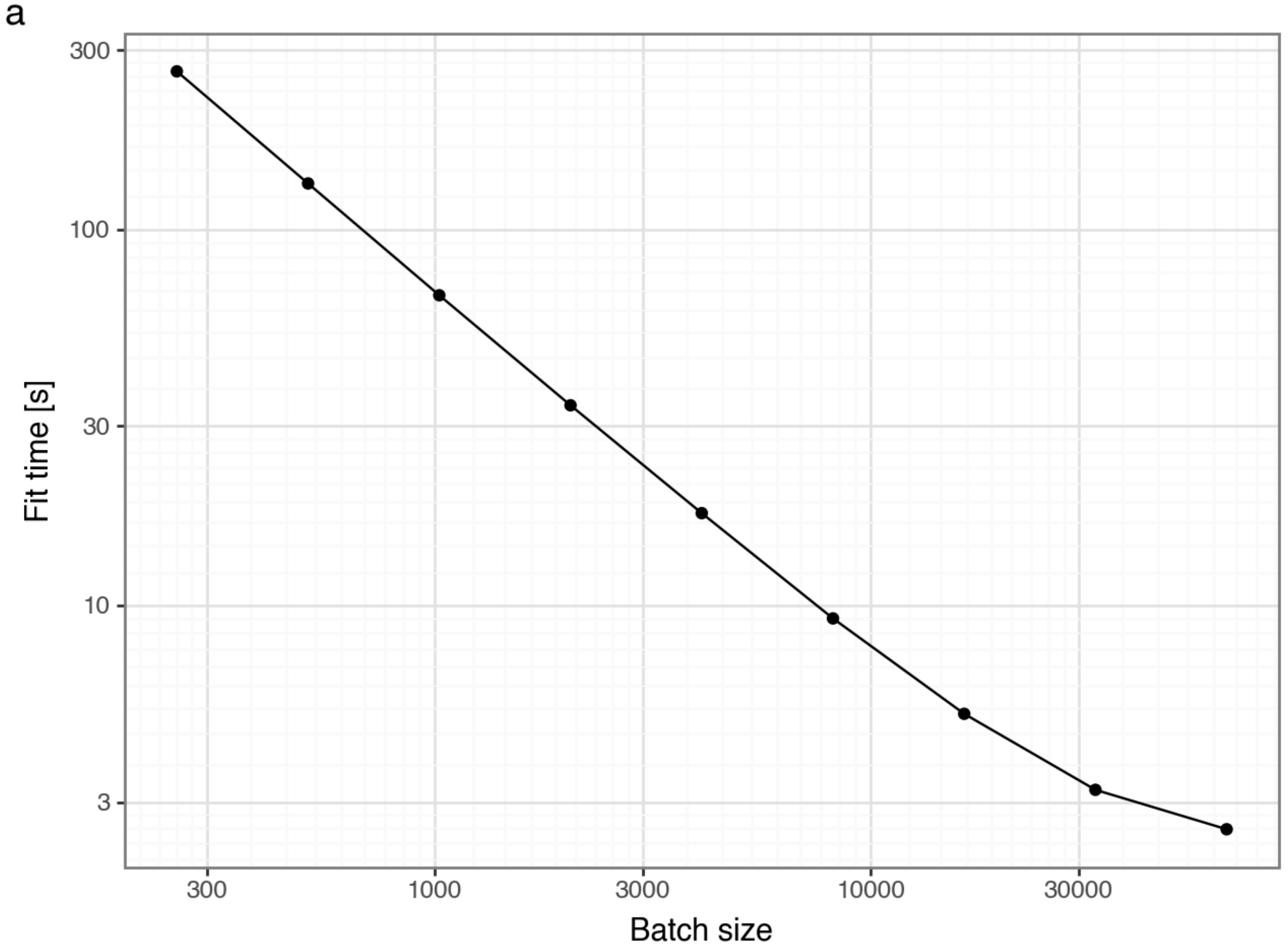